\documentclass{egpubl}
\usepackage{eg2025}
 
\ConferencePaper        
\CGFStandardLicense

\usepackage[T1]{fontenc}
\usepackage{dfadobe} 
\usepackage{amsmath}
\usepackage{amssymb} 

\usepackage{cite}  
\BibtexOrBiblatex
\electronicVersion
\PrintedOrElectronic
\ifpdf \usepackage[pdftex]{graphicx} \pdfcompresslevel=9
\else \usepackage[dvips]{graphicx} \fi

\usepackage{egweblnk} 

\title {TextIM: Part-aware Interactive Motion Synthesis from Text}

\author[Siyuan Fan \& Bo Du \& Xiantao Cai \& Bo Peng \& Longling Sun]
{\parbox{\textwidth}{\centering Siyuan Fan$^{1,2,3}$\orcid{0009-0000-0908-3293}
        and Bo Du\thanks{corresponding author}$^{1,2,3}$\orcid{0000-0001-8104-3448} and Xiantao Cai\thanks{corresponding author}$^{1,2,3}$\orcid{0000-0001-2345-6789} and Bo Peng$^{1,2,3}$\orcid{0000-0001-2345-6789} and Longling Sun$^{1,2,3}$\orcid{0000-0001-2345-6789} }
        \\
{\parbox{\textwidth}{\centering
         $^1$National Engineering Research Center for Multimedia Software, School of Computer Science, Wuhan University, China\\
         $^2$Institute of Artificial Intelligence, Wuhan University, China\\
         $^3$Hubei Key Laboratory of Multimedia and Network Communication Engineering, Wuhan University, China
       }
}
}

\begin{document}
\maketitle

\begin{figure*}[ht]
  \centering
  \includegraphics[width=0.9\linewidth]{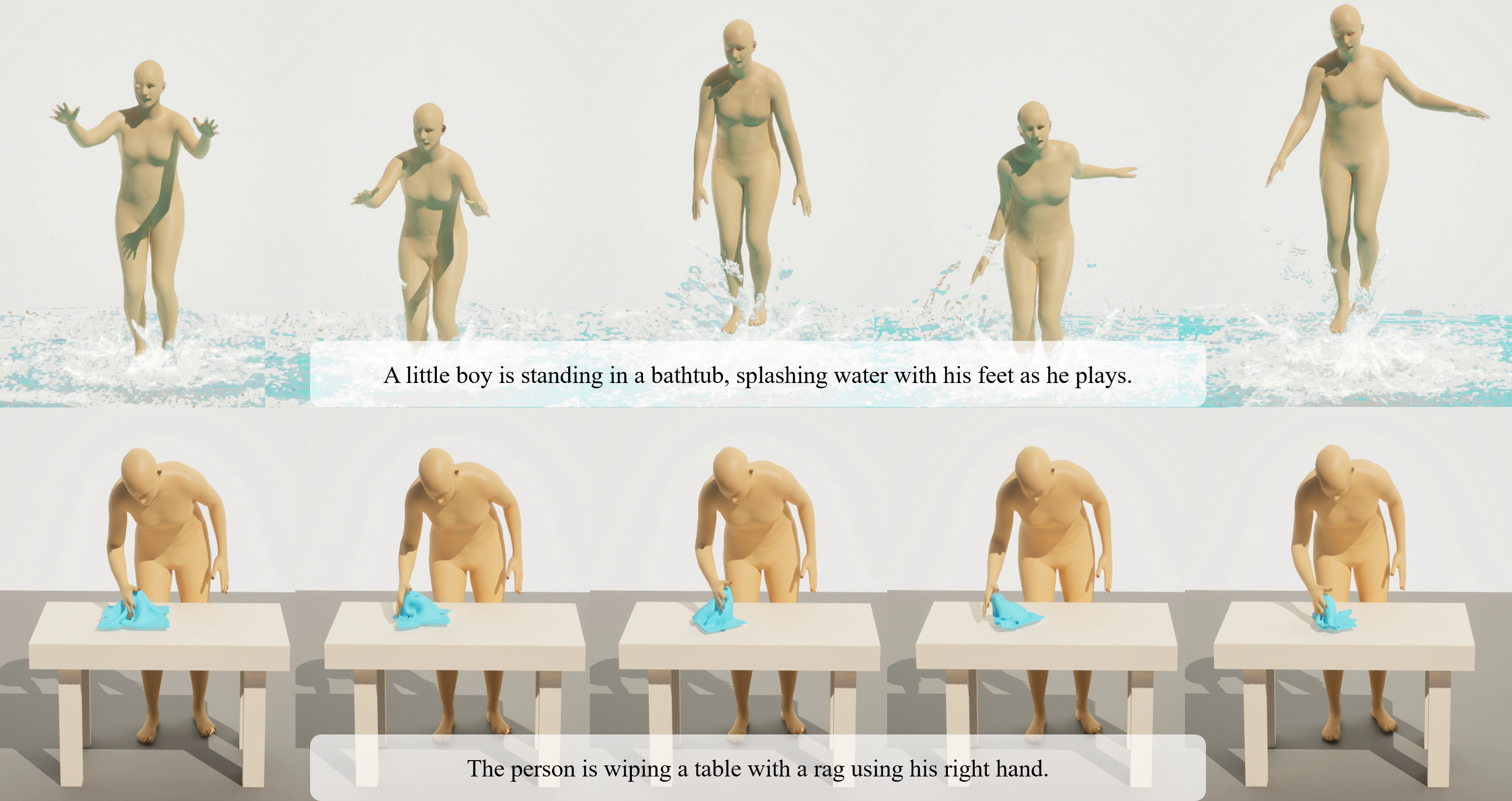}
  \caption{\label{fig:firstExample}
           TextIM generates human interactive motions with part-level semantic accuracy from textual descriptions.}
\vspace{-10pt}
\end{figure*}

\begin{abstract}
In this work, we propose TextIM, a novel framework for synthesizing \textbf{TEXT}-driven human \textbf{I}nteractive \textbf{M}otions, with a focus on the precise alignment of part-level semantics. Existing methods often overlook the critical roles of interactive body parts and fail to adequately capture and align part-level semantics, resulting in inaccuracies and even erroneous movement outcomes. To address these issues, TextIM utilizes a decoupled conditional diffusion framework to enhance the detailed alignment between interactive movements and corresponding semantic intents from textual descriptions. Our approach leverages large language models, functioning as a human brain, to identify interacting human body parts and to comprehend interaction semantics to generate complicated and subtle interactive motion. Guided by the refined movements of the interacting parts, TextIM further extends these movements into a coherent whole-body motion. We design a spatial coherence module to complement the entire body movements while maintaining consistency and harmony across body parts using a part graph convolutional network. For training and evaluation, we carefully selected and re-labeled interactive motions from HUMANML3D to develop a specialized dataset. Experimental results demonstrate that TextIM produces semantically accurate human interactive motions, significantly enhancing the realism and applicability of synthesized interactive motions in diverse scenarios, even including interactions with deformable and dynamically changing objects.


\end{abstract}  
\section{Introduction}

The field of computer animation has increasingly embraced the challenge of text-to-motion synthesis, where faithful and plausible human movements are expected to be generated from descriptive natural language. This task not only bridges a gap between natural language understanding and 3D human motion representation but also enhances applications in virtual reality, gaming, and digital content creation.

While significant advancements have been made in this field, most existing text-to-motion synthesis methods mainly focus on generating human motions that are isolated from the surrounding environment. However, the dynamic nature of most human activities in the real world involves interactions with objects, environments, or other people. Meanwhile, no existing work has sufficiently addressed the challenge of generating motions that accurately align with the semantics of interaction. Recognizing this gap, our research refines the text-to-motion task to specifically address the generation of human interactive motions from textual descriptions. These human interactive motions require sophisticated coordination of specific body parts, posing significant challenges in matching text semantics with detailed interaction movements of individual parts. Our goal is to enhance part-level alignment between interactive movements and the interaction semantics in the text description, concentrating on interactive body parts.

This capability is crucial in applications such as virtual reality, advanced simulations, and interactive training systems, where users interact with virtual environments or avatars that are not physically present. Human interactive motions involve complex movements that require precise control of interactive body parts and comprehensive coordination among all body parts. While interaction movements constitute only a minor part of the overall motion sequence in terms of duration and spatial extent, they remain crucial for the semantically accuracy of the generation condition. The main challenge lies in accurately aligning the interactive semantics described in text with the subtle yet essential interactive movements.

Current text-to-motion methods often fail to precisely align detailed movements and their respective segments of description in the text. They commonly treat the human body as a uniform entity, overlooking the unique contributions of individual body parts in dynamic interactions. This oversight results in a dependency on data that can misinterpret directives such as 'a person wipes the table with their left hand,' where the model might erroneously generate a right-handed action due to the predominance of right-handed data. This misalignment derives from a traditional focus on general correctness rather than a detailed analysis of motion and text, leading to a loss of detailed alignment between the text and the motion. Consequently, the model only generates expected motion sequences when given the text that closely resembles or replicates the form and order of previously learned semantics. In scenarios where sub-actions within the dataset's sentences appear in isolation or are recombined, the model fails to accurately infer the intended actions.

In sight of this, we propose TextIM, a part-aware human interactive motion synthesis model that prioritizes the semantic accuracy of body parts involved in interactions. Specifically, TextIM leverages large language models in the interaction-aware module to extract and comprehensive key interaction instructions and employs a decoupling approach to generate the interactive and non-interactive parts to ensure the accuracy of the interaction. The decoupled approach is designed to parse and understand interactions at a granular level, focusing on the specific body parts most relevant to each described action. This approach also provides a more flexible generation capability, allowing for the synthesis of mixed motion involving both interactive and non-interactive movements, enabling interactions involving multiple parts. While effective at enforcing interaction instruction, the decoupling approach sometimes leads to coordination and spatial continuity issues among the various body parts. To tackle these challenges, drawing inspiration from the action recognition framework ST-GCN, we introduce a spatial coherence module that utilizes graph structure to extract spatial features. These features co-guide the generation of non-interactive parts along with the generated interactive parts motion, ensuring spatial consistency and coherence. This part-aware synthesis approach not only aligns generated motions more closely with the semantics of the text but also enhances the fidelity and realism of the resulting sequences, ensuring that even minor interactive actions are captured with the necessary detail and precision.

To train and evaluate our method, we select and relabel a small-scale human interactive motion dataset based on HUMANML3D, consisting of 1.5K sequences of human motions under categories of daily activities and exercises. To fulfill the motion category, one-third of the motion sequences are non-interactive motions involving basic human actions. Experimental results demonstrate that our approach produces human interactive motions with part-level semantic consistency that aligns with the textual description. We quantitatively evaluate our synthesized sequences on established metrics in the field of motion synthesis and interactive motion synthesis. Further, we conduct a visual perceptual study for human evaluation compared to the related baseline method. 

To summarize, our contributions are as follows:
\begin{itemize}
    \item We present TextIM, a novel diffusion-based approach to generate semantically accurate human interactive motion.
    \item TextIM leverages large language models to comprehensive interactive motion description and generate the interactive motion in a decoupled manner that enhance the detailed alignment between the text and the motion.
    \item TextIM introduces a graph structure to the decoupling approach to maintain the consistency and coherence of the generated human interactive motion sequence, providing enhanced control over the entire motion on top of part-aware movements.
\end{itemize}

\section{Related Work}

\subsection{Text-to-Motion Generation}

Text-to-motion generation can be broadly divided into two categories: action class guided motion synthesis and text instruction guided motion synthesis.

Early studies focus on generating human motion from limited action classes in text form. A substantial amount of works on action2motion datasets has established to solve this task, including UESTC\cite{ji2018large}, NTU-RGB+D\cite{liu2019ntu}, HumanAct12\cite{guo2020action2motion}, and BABEL\cite{punnakkal2021babel}, which have been instrumental in advancing this domain. Generative Adversarial Network (GAN)-based approaches\cite{yu2020structure, degardin2022generative} enhance the generation ability of GAN architecture with Graph Convolutional Network (GCN) to generate accurate motion sequences. Variational Autoencoder (VAE)-based methods\cite{guo2020action2motion, petrovich2021action, lu2022action, lucas2022posegpt, lee2023multiact} leverage an encoder-decoder structure to effectively generate motion sequence, reflecting the evolution in the field's technical approach.

However, the limited predefined action classes is constrained to illuminate human action willing, leading to the exploration of text instruction guided motion synthesis. KIT\cite{plappert2016kit} and HUMANML3D\cite{guo2022generating} are the two main dataset used in text instruction guided motion synthesis, consisting of paired motion sequences and corresponding text annotation. VAE-based methods\cite{guo2022generating, petrovich2022temos, guo2022tm2t, athanasiou2022teach, zhang2023t2m, lin2023being} tends to learn the mapping relation between textual descriptions and human motion in a joint embedding space. Additionally, integrating the pretrained CLIP model\cite{radford2021learning} into the text-to-motion generation process\cite{tevet2022motionclip, hong2022avatarclip} introduces prior cross-modal knowledge, thereby enhancing semantic comprehension capabilities. More recent approaches\cite{tevet2022human, zhang2024motiondiffuse, shafir2023human, chen2023executing, yuan2023physdiff, jiang2024motiongpt} adopt diffusion model as the backbone network, which offers robust frameworks to accept multiple conditioning types within the same model. 

Despite these advances, the demands of generating interactive human motions based on detailed textual instructions have not been fully explored. Interactive motion synthesis presents unique challenges, particularly in capturing the subtleties of context-dependent interactions that go beyond general motions. The complexity of accurately synthesizing interactive motion requires a deeper semantic understanding and a more granular approach to motion generation.

\subsection{Interaction Motion Generation}

Interaction motion generation mainly includes human-scene interaction (HSI) task and human-object interaction (HOI) task. Since recent human-scene interaction studies\cite{hassan2021stochastic, wang2021synthesizing, wang2022towards, mir2023generating} mostly involving navigation and obstacle avoidance goal, we focus on more related HOI task.

Early HOI synthesis works\cite{taheri2022goal, zhao2022compositional, kulkarni2023nifty} start with reducing the distance between human and static objects while maintaining a correct contacting pose. Subsequent studies\cite{wu2022saga} extend human interaction to dynamic objects, yet they remained limited to small objects and only considering interactions involving hands. Meanwhile, human intention represented in the form of action class-object pairs\cite{ghosh2023imos, li2024task} are involved to the generation guidance. Recent approaches expand object information such as geometry\cite{li2023controllable, peng2023hoi}, trajectory\cite{li2023object} and contact\cite{diller2023cg, ma2024contact} to advance HOI. However, comprehensive dataset for HOI that involves whole-body human motion sequences along with corresponding human information and object information is still vacant. Existing HOI datasets\cite{bhatnagar2022behave, li2023object} lacks free-form language description of human intention.

Although text-conditioned interactive motion synthesis and HOI tasks operate within similar interactive scenarios, our research points out certain challenges in the traditional HOI approach. HOI studies primarily focus on the physical realism of interactions between humans and objects, emphasizing detailed object information such as geometries and contacts. Consequently, HOI methods may find it challenging to handle interactions with soft or deformable objects, whose shapes and contacts change during the interaction process. In contrast, TextIM concentrates on the semantic consistency between human motions and textual descriptions from a human-centric perspective, enabling more flexible handling of a wider variety of interactions. Furthermore, TextIM can be extended to improve semantic guidance within the HOI domain, thereby increasing the accuracy and applicability of interaction-driven motion synthesis across diverse scenarios.


\begin{figure*}[ht]
  \centering
  \includegraphics[width=\linewidth]{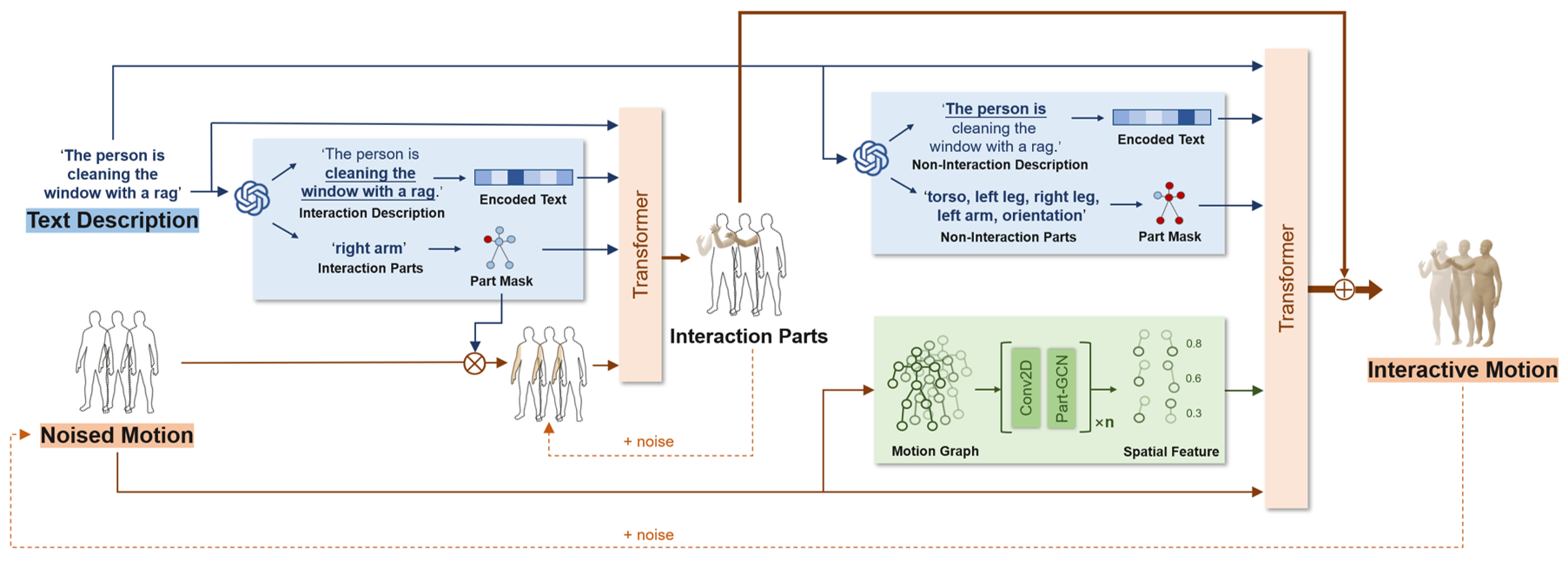}
  \caption{\label{fig:network}
   \textbf{TextIM overview.} TextIM synthesizes human interactive motions in a decoupled manner to align part-level motions with textual semantics. Given a textual instruction, we employ LLM to extract interaction instructions and body parts to generate the corresponding interaction motion. Subsequently, we use GCN to learn spatial features based on interaction motion to guide the generation of the final result along with the interaction information.}
\vspace{-10pt}
\end{figure*}

\section{Method}

\subsection{Overview}
Our proposed TextIM architecture is shown in Figure \ref{fig:network}. Given a conditional textual description $c$, our goal is to generate human interactive motion sequences $x$ that are semantically accurate at part-level. To tackle the misalignment between part-level texts and motions, TextIM employs a decoupled generation pipeline based on a diffusion framework to deal with human interactive motions. By decomposing both texts and motions into fine-grained representations, TextIM leverages a two-module system to fulfill the generation of the final motion sequence. The interaction-aware module is designed to encode the interactive descriptions accurately, allowing for the precise generation of fine-grained interactive motions. However, naively segmenting the motion into distinct body parts results in unnatural outcomes in the generated motion sequences. Consequently, we introduce the spatial coherence module to ensure coordination within the different body parts, adjusting the potential inconsistencies caused by independent part-wise motion generation. 

In the following, we first present the motion representation and the general diffusion generation model in Section 3.1. We then give the composition of the interaction-aware module in Section 3.2. Last, we elucidate the details of the spatial coherence module in Section 3.3.

\subsection{Human Motion Diffusion Model}

We begin our introduction with a brief overview of the fundamentals of the human motion generation task, motion representation, and the conditional diffusion model.

Human interactive motion $x \in \mathbb{R}^{T \times D}$ consists of a sequences of human poses $p$, where $T$ represents the number of frames in the motion sequence and $D$ represents the dimensions characterizing the human pose. Each human pose $p=\left[r^a, r^l, r^h, p, v, r, f\right]$ includes the angular velocity $r^a$, the linear velocities $r^l$, the height $r^h$ of root joint in the SMPL human skeleton structure and the relative positions $p$, the relative velocities $v$, the relative rotations $r$ of each joint compared to the root joint and the binary feature $f$ of foot contacts with the ground.

Diffusion model shows extraordinary ability in motion generation tasks. In this paper, we leverage conditional diffusion model in a classifier-free manner to generate interactive motion from text description. In the forward process, diffusion model gradually adds noise to the raw motion data $x_0$ using a Markov chain.
\begin{equation}\label{eq:3_noise}
q\left(\mathbf{x}_t \mid \mathbf{x}_{t-1}\right)=\mathcal{N}\left(\mathbf{x}_t ; \sqrt{\alpha_t} \mathbf{x}_{t-1}, \left(1-\alpha_t\right) \mathbf{I}\right)
\end{equation}
where $x_t$ represents the motion sequence at the $t^{th}$ noise step, $\alpha_t=1-\beta_t$ and $\beta_t$ represents a variance schedule.

In the reverse process, conditional diffusion model learns to generate clean data from pure Gaussian noise $x_t$ with condition $c$.
\begin{equation}\label{eq:3_denoise}
p_\theta\left(\mathbf{x}_{t-1} \mid \mathbf{x}_t, \mathbf{c}\right)=\mathcal{N}\left(\mathbf{x}_{t-1} ; \mu_\theta\left(\mathbf{x}_t, t, \mathbf{c}\right),\left(1-\alpha_t\right) \mathbf{I}\right)
\end{equation}
where $\mu_\theta$ represents the learned mean. Instead of predicting the noise $\epsilon_t$ at the $t^{th}$ noise step, we follow Telvet et al. \cite{tevet2022human} and predict the clean motion $\hat{\mathbf{x}}_0=f_\theta\left(\mathbf{x}_t, t, \mathbf{c}\right)$, so mean $\mu_\theta$ is formulated as 
\begin{equation}\label{eq:3_mu}
\mu_\theta\left(\mathbf{x}_t, t, \mathbf{c}\right)=\frac{1}{\sqrt{\alpha_t}}\left(\mathbf{x}_t-\frac{\beta_t}{\sqrt{1-\bar{\alpha}_t}}\left(\mathbf{x}_t-\sqrt{\bar{\alpha}_t} \hat{\mathbf{x}}_0\right)\right)
\end{equation}
where $\bar{\alpha}_t=\prod_{i=0}^t \alpha_i$. The model parameters $\theta$ are optimized by minimizing $\left\|\hat{x}_0-x_0\right\|_2^2$.

\subsection{Interaction-aware Module}

When textual instruction is given, fully and accurately comprehending the semantic information is key to generating plausible results. In this section, we will introduce our interaction-aware module to fulfill such an objective.

\textbf{Interaction semantic extraction.} In human interactions, the specific interactive motion typically has the nature of being small-scale and detailed compared to the entire movement, yet they possess the greatest significance. For instance, in the action of walking through the corridor to pick up a file, the act of picking occurs over a brief time and within a limited spatial scale, while it carries significantly more importance than the more extended and broader movement of walking. Therefore, using text instructions directly as the condition for a diffusion model can lead to the oversight of crucial interaction information. To address this, we employ a large language model to identify and extract the crucial contents of interaction semantics at the body part level from the original instruction.

To leverage the semantic understanding capabilities of large language models, we use OpenAI's GPT-3.5\cite{chen2021evaluating} to extract interaction instruction and corresponding body parts from the given textual instruction. To be more specific, we structure the request prompt from four key aspects. (i) First, we clarify the objective by specifying the expected outcome, with the prompt directly stating: "Identify all possible body parts involved in interacting with an object from the given sentence and extract the exact phrase that describes their action." (ii) Next, we give a precise definition of interaction, as its meaning varies under different circumstances. "An 'interaction' involves any purposeful physical engagement with an object, such as holding, touching, lifting, carrying, moving, manipulating, or using it in any way." (iii) Moreover, we provide optional answer choices to avoid responses that are overly fine-grained or excessively coarse-grained. "If no body parts are interacting with an object, respond with 'none'. Choose body parts from: left arm, right arm, left leg, right leg, torso, pelvis." (iv) Last, few-shot examples of question-answer pairs are given for guidance, along with the actual question. "Here are examples of how to format your responses: examples. Provide your answer for the following sentence: [QUESTION]."

Based on the aforementioned prompt, the responses from GPT3.5 are usually appropriate as expected. However, to ensure all responses meet the standard, format checking process and content amendment process are designed to avoid unnecessary costs. In rare cases, responses may include unexpected interaction body parts or incorrect extraction of interaction instructions. The interactive body parts in the response are primarily verified against the expected body part choices. Any unexpected body parts are transformed into corresponding optional parts based on a predetermined mapping. The predetermined mapping based on SMPL skeletal hierarchy maps common body parts to the six optional parts. For example, "hip, knee, ankle, foot" are mapped to "foot" and "collarbone, shoulder, elbow, wrist"  are mapped to "arm" and so on. Subsequently, the text format of the response is checked to enable information extraction for the following post-processes. Each question is given three attempts to conform to the expected answer. If the answer consistently fails to pass the verification, it is set to the default 'none' type. 

\textbf{interactive motion segmentation.} During the training and generating processes, interacting body parts and their corresponding interaction instructions are extracted from the initial text description. By focusing on part-level decomposition of the text, TextIM enables the precise generation of interaction-specific motions, addressing the challenge of aligning detailed text descriptions with the corresponding motion at a granular level. The decomposition approach ensures that each body part’s movement is both contextually relevant and semantically consistent, significantly enhancing the fidelity and realism of the generated interactive motions.

Since interactive parts in a whole-body motion are subtle and minor, we decouple them from the rest parts to better learn interaction semantics. The extracted body parts are converted into a binary mask $m \in \mathbb{R}^{263}$ according to the compounded motion representation in HUMANML3D\cite{guo2022generating}. The binary mask $m$ is then applied to the original whole-body motion $x \in \mathbb{R}^{T \times D}$ for the corresponding interaction body parts $x_{inter} \in \mathbb{R}^{T \times B}$, where $B$ represents the corresponding interaction parts dimension. This approach not only encourages the model to focus on fine-grained interactive motion features during the first generation stage but also supports the inclusion of arbitrary interaction parts within a single motion.

The binary part mask is processed through a learnable linear projection, which is then utilized as guidance for generation. Meanwhile, the interaction instructions are encoded through the pre-trained CLIP model as text instruction to serve as supplementary guidance alongside the masks. Together, the original textual instruction, the transformed body part mask, and the encoded interaction instructions guide the generation of interactive motions within the interaction-aware module.

\begin{table*}[t!]
\centering
\caption{Quantitative evaluation on relabelled interactive HUMANML3D dataset.}
\label{tab:metrics}
\begin{tabular}{lcccccc}
\hline
Method & \begin{tabular}[c]{@{}c@{}}R Precision\\ (top3)↑\end{tabular} & Multimodal Dist↓     & MPJPE↓               & MPVPE↓               & Hand JPE↓            & Foot JPE↓            \\ \hline
MDM    & $0.203^{\pm 0.012}$                                                   & $6.818^{\pm 0.075}$          & $1.296^{\pm 0.019}$          & $7.367^{\pm 0.081}$          & $2.450^{\pm 0.029}$          & $1.294^{\pm 0.021}$         \\
TextIM   & $0.220^{\pm 0.012}$                                                   & $6.476^{\pm 0.058}$          & $1.144^{\pm 0.011}$          & $5.768^{\pm 0.036}$          & $2.017^{\pm 0.017}$          & $1.147^{\pm 0.015}$          \\ \hline
       & \multicolumn{1}{l}{}                                          & \multicolumn{1}{l}{} & \multicolumn{1}{l}{} & \multicolumn{1}{l}{} & \multicolumn{1}{l}{} & \multicolumn{1}{l}{}
\end{tabular}
\vspace{-10pt}
\end{table*}

\subsection{Spatial Coherence Module}

Once the interactive parts of an interactive motion are confirmed, we proceed to complement the entire body movement based on the generated parts. In this section, we introduce the spatial coherence module to accomplish the entire generation process while maintaining consistency and harmony throughout the full-body movement.

\textbf{Spatial feature extraction.} Similar to the interactive body parts, the entire body motion is also generated within a conditional diffusion model. However, the generation of the remaining motion is influenced by the interactive motion generated by the interaction-aware module and must maintain spatial coherence with it. Consequently, we extract spatial features using a Part Graph Convolutional Network (Part-GCN) during the training process to ensure spatial coherence and consistency throughout the motion sequence.

The graph $G=(V, E)$ is constructed based on the motion sequence $x_0$, encapsulating both temporal and spatial dynamics. The node set $V=\left\{v_{t i} \mid t=1, \ldots, T, i=1, \ldots, N\right\}$ contains nodes for $N$ joints across $T$ frames, which collectively represent the human motion sequence. The feature map $F(v_{ti})$ of each node consists of the joint's rotation and position information in the motion data. The edge set $E = \left\{(v_{ti}, v_{tj}) \mid (i, j) \in C\right\} \cup \left\{(v_{ti}, v_{t+1,i}) \mid t=1, \ldots, T, i=1, \ldots, N\right\}$, where $C$ represents the anatomical connections between human joints. This set includes edges representing both intra-frame joint connections based on real human anatomy and inter-frame temporal connections that illustrate the trajectory of joints.

Feature aggregation within the Part-GCN utilized for spatial feature extraction is formalized as follows:
\begin{equation}\label{eq:3_fout}
F^{(l+1)}=\sigma\left(\sum_{k=1}^k A_k F^{(l)} W_k^{(l)}\right)
\end{equation}
where $F^{(l+1)}$ represents the output feature matrix of layer $l$, $\sigma$ is the non-linear activation function ReLU, $A_k$ is the adjacency matrices, and $W_k^{(l)}$ is the learnable weight matrices corresponding to each adjacency matrix at each layer $l$. Unlike traditional Graph Convolutional Networks, Part-GCN leverage multi-layer adjacency matrices that represent practical connection relationships, enabling feature aggregation tailored to the partition of human body parts. Furthermore, learnable weight matrices are involved to assess the varying importance of different parts and the scale of convolutional layers.

\begin{figure}[ht!]
  \centering
  \includegraphics[width=0.8\linewidth]{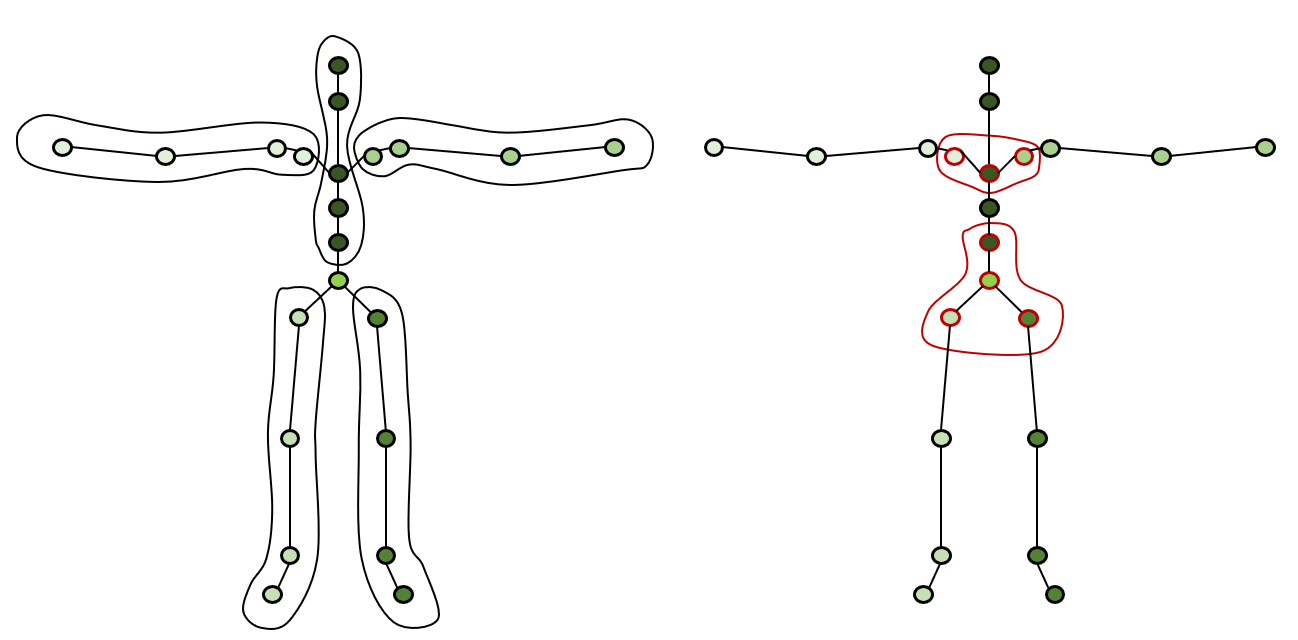}
  \caption{\label{fig:partition}
  Multi-layer adjacency matrices enables intra-part and inter-part feature aggregation. The black loops show intra-part connections of each body parts and the red loop shows the inter-part connection.}
\end{figure}

The joint nodes are segmented into $k$ subsets based on the human skeletal structure and body part partitions,  allowing for the aggregation of joint feature intra-part and inter-part. As illustrated in Figure \ref{fig:partition}, the human body is naturally divided into five parts: left arm, right arm, torso, left leg, and right leg. Nodes from different body parts which directly connects to each other form the subset for inter-part information updating.

Human motion sequences are processed through 2D convolutional layers to compress temporal dimension and through part graph convolution layers to extract spatial features. After feature aggression and information updating, joint nodes connectivity is assessed by calculating cosine similarity among node feature vectors $S = F_{\text{norm}}^{(l)} \cdot (F_{\text{norm}}^{(l)})^T$, where $F_{\text{norm}}^{(l)} = \frac{F^{(l)}}{\sqrt{\sum (F_{\text{joint}}^{(l)})^2}}$. This spatial similarity feature is subsequently utilized to guide the generation of whole-body motions, ensuring spatial coherence.

\textbf{Interaction guided generation.} To integrate the interaction body parts into the final result, we overwrite the corresponding interaction body parts onto the generated result in the spatial coherence module after each denoising step. The modified generation result is composed of the interactive motion and the newly generated motion based on the body part mask $m$, defined as:
\begin{equation}\label{eq:3_label_single}
\hat{x}_0^{\prime}=m \odot x_{\text {inter }}+(1-m) \odot \hat{x}_0
\end{equation}
We also modify the optimization objective to $\left\|\hat{x}_0^{\prime}-x_0\right\|_2^2$ to direct the optimizing focus towards the remaining parts of the motion.
 
Furthermore, to keep consistent with the interaction-aware module, we maintain similar condition types to guide the motion generation in the spatial coherence module. The body mask $m^{\prime}$ is reversed from the original mask $m^{\prime}=1 - m$, representing the opposite active body parts. The non-interaction instructions, which refer to the remaining parts of the original text that are not covered by the extracted interaction instructions, are used as an additional condition to enhance the semantic guidance for motion generation. Together, the original textual instruction, the transformed body part mask, the encoded non-interaction instructions, and the spatial feature guide the generation of the final result motion within the spatial coherence module.

\section{Experiments}

\begin{figure*}[ht!]
  \centering
  \includegraphics[width=\linewidth]{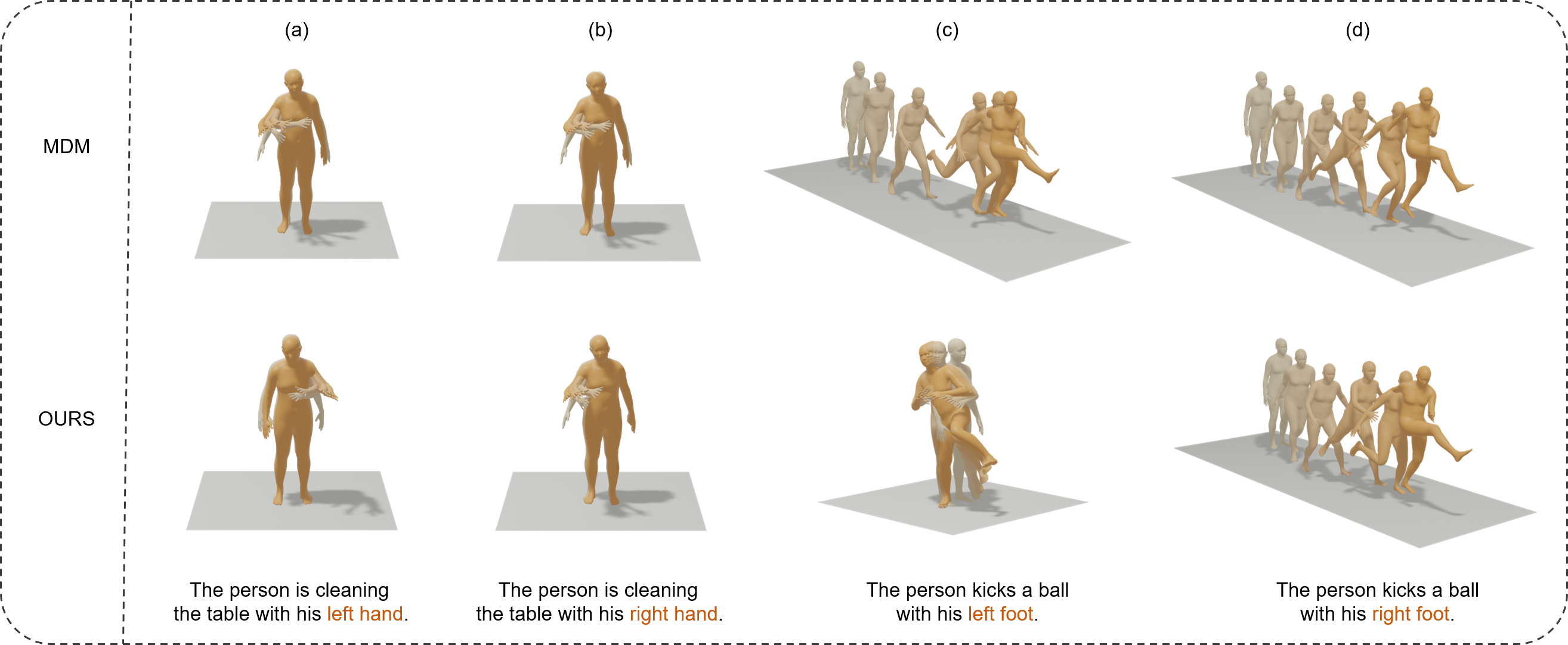}
  \caption{\label{fig:leftright}
  Semantic accuracy. TextIM accurately generates human motions based on detailed, part-level textual descriptions, ensuring semantic precision in interactive motion generation. The examples illustrate how specifying an interactive body part for the same action can lead to different motion outcomes.}
\end{figure*}
\begin{figure*}[ht!]
  \centering
  \includegraphics[width=\linewidth]{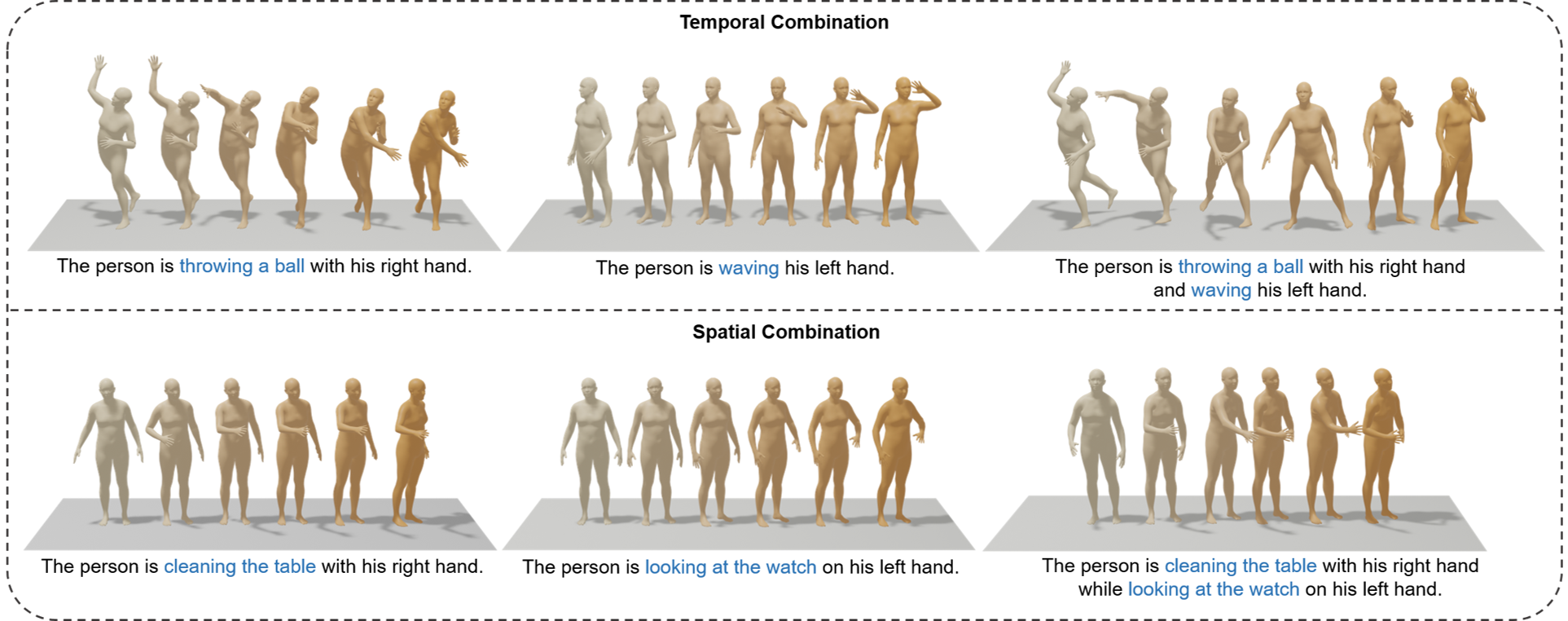}
  \caption{\label{fig:combination_row}
   Motion combination. TextIM combines existing motions into new ones across both temporal and spatial dimensions, enhancing the data efficiency of interactive motion generation. The examples illustrate how motions can be combined both temporally and spatially to produce varied outcomes.}
\end{figure*}

\subsection{4.1 Dataset}
Since no dataset exists is designed for text-driven interactive motion synthesis, we selected 992 interactive motion sequences from HUMANML3D dataset\cite{guo2022generating} under the categories of daily activities and sports, based on the original text descriptions and the animations of the motion. Subsequently, we manually annotated each motion sequence with three text instructions that describe the human motion in various ways, ranging from detailed to sketchy. These annotated captions are then processed through a standard word segmentation procedure. To broaden motion range of motions and to include fundamental actions absent from the available interactive motion sequences, such as "run" and "turn around", we randomly selected an additional 500 motions and their original text annotation from the unselected portions of daily activities and sports category to enrich our dataset. All motions represent human movement across 22 body joints and have an average duration of 7 seconds, resulting in a total motion length of approximately 2 hours. In the end, our dataset comprises 1492 3D human interactive motion sequences along with textual descriptions to train and evaluate our method.

\subsection{4.2 Evaluation Metrics}
We employ established metrics from both text-to-motion generation task and human-object interaction generation task to evaluate our synthesized results. Specifically, we utilize evaluation metrics from the text-to-motion generation task to evaluate the condition matching quality, while metrics from the human-object interaction generation task are applied to evaluate the quality of human interactive motion results.
 
First, we adapted general evaluation metrics from text-to-motion generation tasks\cite{guo2022generating} to better align with the objectives of interactive motion synthesis.
Given that human interactive motion synthesis primarily focuses on interaction movements, we have increased the semantic significance of interaction by replacing the standard text embedding used in R-Precision and MM-dist metrics with interaction text embedding. Furthermore, we omit the FID and Multimodality metrics, as they calculate the entire motion embedding and do not reflect the crucial yet minimal feature of interactive motion. Instead, we leave motion quality evaluation to metrics in the human-object interaction generation task.

As for metrics used in human-object interaction generation task\cite{li2023object}, we select the metrics that best fit our current task. These include mean per-joint position error (MPJPE) and mean per-joint velocity error (MPVPE). Additionally, we choose hand joint position error (HandJPE) and foot joint position error (FootJPE), as these body parts are most involved in interactive motions.

\subsection{4.3 Results}

\textbf{Baselines.} Our work focuses on generating human interactive motion from textual descriptions. Given that existing text-to-motion methods have different focuses and objectives, we select the prominent MDM method \cite{tevet2022human} to represent these methods for our comparative analysis.

\textbf{Quantitative results.} Table \ref{tab:metrics} reports the quantitative results on the relabeled interaction HUMANML3D dataset. Compared to the baseline method, TextIM has a better performance on all metrics, indicating its effectiveness in synthesis human interactive motions.

\textbf{Qualitative results.} To illustrate the performance of TextIM, we present qualitative evaluation results that demonstrate its capacity for fine-grained semantic accuracy. As shown in Figure \ref{fig:leftright}, the baseline method MDM fails to follow the textual instructions specifying body parts, generating similar results for two distinct inputs. TextIM accurately follows the semantic nuances of description such as "clean the table with the left hand" and "kick a ball with the right leg" showcasing its robust ability to distinguish and apply part-level interactions as directed by the input description.

Figure \ref{fig:combination_row} illustrates the capability of TextIM to automatically combine existing motions from the dataset into new, contextually relevant motions across spatial or temporal dimensions. For example, the motion labeled "throwing a ball and waving" successfully integrates these two distinct actions by sequentially connecting the "waving" movements immediately following the "throwing" action. Meanwhile, the motion "cleaning the table while looking at the watch" demonstrates simultaneous integration by combining "table cleaning" with "watch looking," executing both actions concurrently. The decomposition and recombination of interaction semantics allow TextIM to effectively mimic the natural formation of human actions, showcasing its advanced capacity for understanding complex motion dynamics.

\subsection{4.4 Application}
\begin{figure}[h!]
  \centering
  \includegraphics[width=\linewidth]{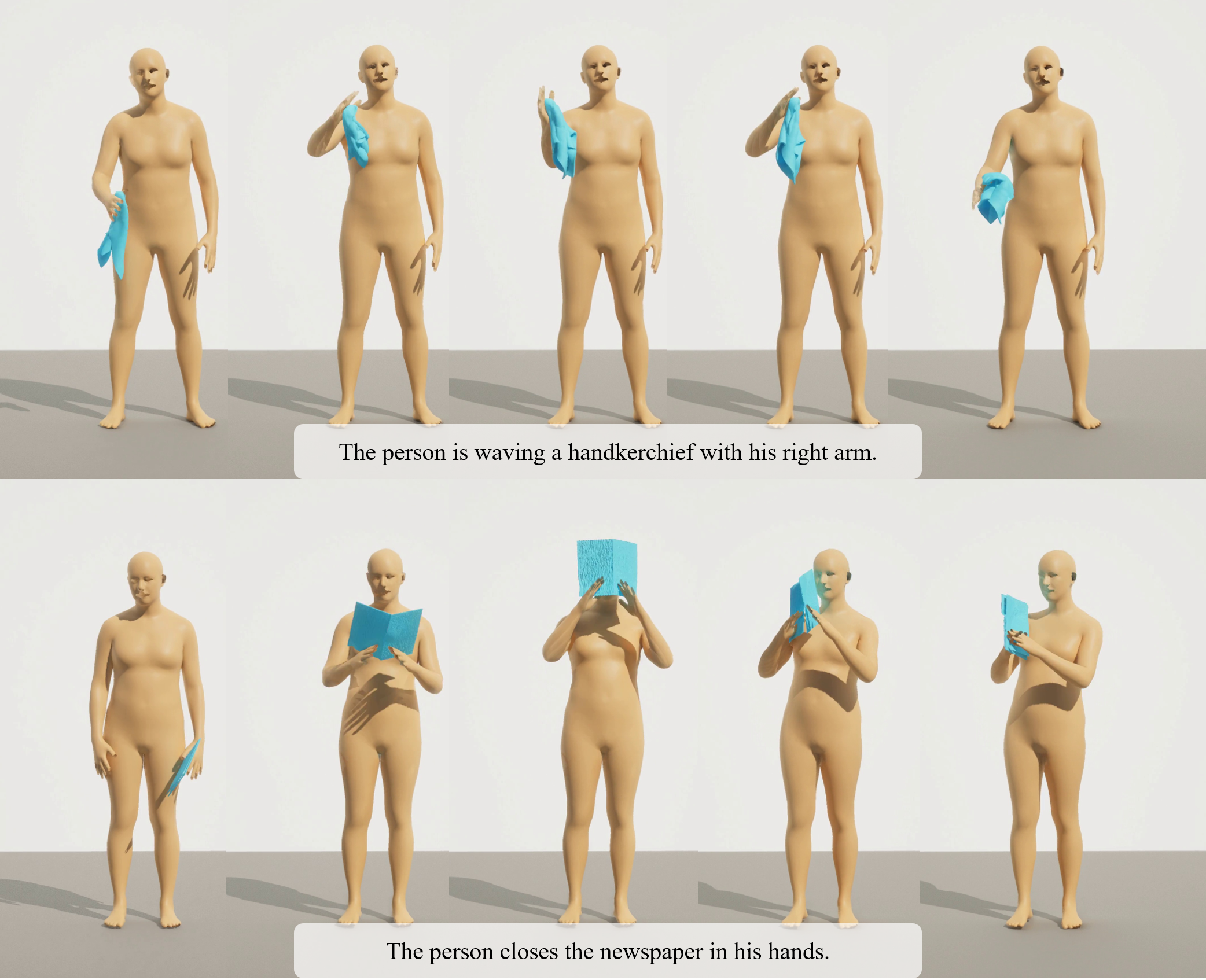}
  \caption{\label{fig:application2}
  Human-object interaction results of "waving a handkerchief" and "closing the newspaper".}
\end{figure}

We present a practical application of TextIM, enabling human interaction synthesis with soft objects and deformable objects. Objects do not have their own motion intentions and must be manipulated in accordance with human movement. In our approach, the motions of objects are not directly generated but are instead derived through the simulated interaction effects alongside the human motions. We utilize the physical simulation capabilities of Unreal Engine to fulfill our goal. By attaching a physically simulated object to the generated human interactive motion within simulation software, we can accurately reproduce complex interaction dynamics. As illustrated in Figure \ref{fig:application2}, we show the interactions like "waving a handkerchief" and "splashing water" which are challenging to achieve with current methods. These examples highlight TextIM's unique ability to integrate human motion synthesis with physical object interactions, providing a robust demonstration of its practical applications in scenarios where human-centric interactions are expected.

\section{Conclusion}

In summary, we proposed TextIM for synthesizing human interactive motion guided by textual descriptions with an emphasis on part-aware interaction alignment. Specifically, we employ a decoupled diffusion framework to prioritize precise alignment between interaction instructions and motion generation. Then a spatial coherence module is employed to extract spatial features of interactive motion to maintain spatial coherence. We selected and relabelled interactive motions in HUMANML3D dataset for evaluation. TextIM achieves better quantitative results on the interaction dataset compared to previous text-to-motion work, specifically excelling in part-level semantic accuracy and effective motion combination ability in complex interactive motion synthesis. Moreover, our results can be directly applied to all kinds of objects, including soft objects and deformable objects using physical simulation environments compared to human-object interaction approaches.

\textbf{Limitations and future work.} Although TextIM introduces a novel approach for generating interactive motion from a human-centric perspective, integrating object information more deeply into the generation process, especially for interactions with rigid objects, remains an area for future enhancement. Meanwhile, the vacancy of interactive motion datasets that include precise finger movements constrained the fidelity of interactive motion synthesis.

\bibliographystyle{eg-alpha-doi} 
\bibliography{egbibsample}

\end{document}